\documentclass[runningheads]{llncs}

\usepackage{eccv}

\usepackage{eccvabbrv}

\usepackage{graphicx}
\usepackage{booktabs}
\usepackage{bm}
\usepackage{multirow}
\usepackage{multicol}
\usepackage{float}
\usepackage{arydshln}
\usepackage{makecell}
\usepackage{bigstrut}
\usepackage{adjustbox}
\usepackage{wrapfig}
\usepackage{graphicx}
\usepackage{array}
\usepackage{caption}
\usepackage{amsmath}
\usepackage[accsupp]{axessibility}  
\newcommand*{\menlo}{\fontfamily{lmtt}\fontsize{9}{9}\selectfont }

\newcommand{\myparagraph}[1]{\vspace{0pt} \noindent \textbf{#1}}

\newcommand{\reffig}[1]{Figure~\ref{fig:#1}}

\newcommand{\reftbl}[1]{Table~\ref{tbl:#1}}

\newcommand{\lblsec}[1]{\label{sec:#1}}

\newcommand{\ignorethis}[1]{}

\usepackage{hyperref}

\usepackage{orcidlink}

\begin{document}

\title{Customized Generation Reimagined: Fidelity and Editability Harmonized} 


\author{Jian Jin\orcidlink{0009-0005-0466-3121} \and
Yang Shen\orcidlink{0000-0002-6344-9951} \and
Zhenyong Fu\thanks{Corresponding author.} \and
Jian Yang\inst{\star}}

\authorrunning{J. Jin et al.}

\institute{
PCA Lab, Key Lab of Intelligent Perception and Systems for High-Dimensional Information of Ministry of Education, and Jiangsu Key Lab of Image and Video Understanding for Social Security, School of Computer Science and Engineering, Nanjing University of Science and Technology \\
\email{\{jinj,shenyang\_98,z.fu,csjyang\}@njust.edu.cn}}

\maketitle

\begin{abstract}
Customized generation aims to incorporate a novel concept into a pre-trained text-to-image model, enabling new generations of the concept in novel contexts guided by textual prompts.
However, customized generation suffers from an inherent trade-off between concept fidelity and editability, i.e., between precisely modeling the concept and faithfully adhering to the prompts.
Previous methods reluctantly seek a compromise and struggle to achieve both high concept fidelity and ideal prompt alignment simultaneously.
In this paper, we propose a \emph{``\underline{D}ivide, \underline{C}onquer, then \underline{I}ntegrate''} (DCI) framework, which performs a surgical adjustment in the early stage of denoising to liberate the fine-tuned model from the fidelity-editability trade-off at inference.
The two conflicting components in the trade-off are decoupled and individually conquered by two collaborative branches, which are then selectively integrated to preserve high concept fidelity while achieving faithful prompt adherence.
To obtain a better fine-tuned model, we introduce an \emph{\underline{I}mage-specific \underline{C}ontext \underline{O}ptimization} (ICO) strategy for model customization.
ICO replaces manual prompt templates with learnable image-specific contexts, providing an adaptive and precise fine-tuning direction to promote the overall performance.
Extensive experiments demonstrate the effectiveness of our method in reconciling the fidelity-editability trade-off.
Code is available at \href{https://github.com/jinjianRick/DCI_ICO}{https://github.com/jinjianRick/DCI\_ICO}.
\end{abstract}

\section{Introduction}
\label{sec:intro}
The recent large-scale text-to-image models, such as
GLIDE~\cite{ref19}, Imagen~\cite{ref20}, and DALL-E 2~\cite{ref21},
have demenstrated astonishing image rendering ability, guided by text
prompts in natural language.
These models, pretrained on extensive caption-image pairs, allow users
to generate visually realistic images with novel scenes and diverse
styles.
Although text-to-image generation has progressed impressively, implanting new visual concepts into pretrained models remains a major open challenge in this area.
For example, users may wish to customize the models with their own
specific photos.
More personalized or some unusual visual concepts are difficult to
describe merely with text prompts.
In such scenarios, pretrained text-to-image models need to be adapted to incorporate these new visual concepts.

\begin{figure}[!t]
    \centering
    \includegraphics[width=0.85\linewidth]{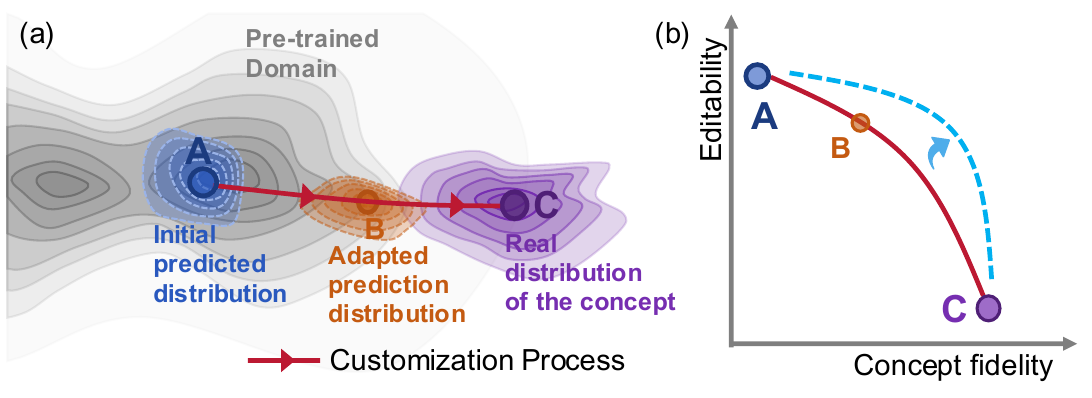}
    \caption{\textbf{Concept fidelity and editability trade-off in customized generation.}  
    (a) is an illustration of the customization process, and (b) shows the corresponding variations in concept fidelity and editability during the customization process.
    \textbf{(a)}: The distribution is portrayed in two dimensions. In the ``pre-trained domain'', deeper colors indicate higher editability.
    \textbf{(b)}: Concept fidelity and editability outline a trade-off curve (red line) in the customization process. Our method (blue dotted line) is designed to free the fine-tuned model from this trade-off, and provide a powerful inference-time adjustment mechanism to achieve satisfactory generations for all query prompts using a single fine-tuned model.
    }
    \label{fig:trade_off}
\end{figure}

Concept customization of text-to-image models, proposed
in~\cite{ref6,ref2,ref3}, aims to implant a specific or personal concept into pretrained models, using a few reference images depicting the custom concept.
In the customization process, the pre-trained model (and a concept token in some works) is adapted to learn the distribution of the new concept, as illustrated in~\Cref{fig:trade_off} (a), binding the concept with the concept token.
Distributions that are closer to the pre-trained domain (\emph{e.g.}, point A) tend to be more easily edited into various contexts but suffer from deficiencies in concept fidelity.
In contrast, distributions closer to the customized domain (\emph{e.g.}, point C) learn the concept well but experience a reduction in editability.
Therefore, the inherent \emph{fidelity-editability} trade-off emerges during the customization process, as shown in~\Cref{fig:trade_off} (b).

\begin{figure*}[!t]
    \centering
    \includegraphics[width=\linewidth]{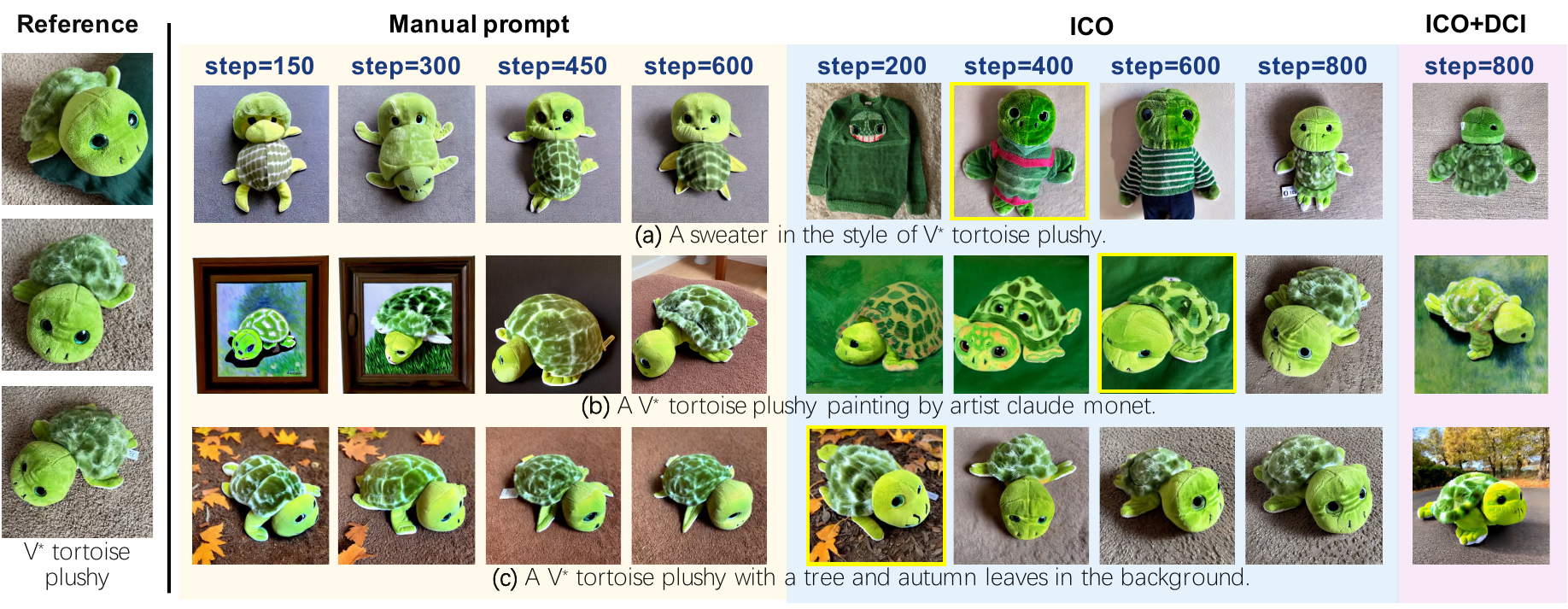}
    \caption{\textbf{Visualization of query examples varying in customization processes.} 
    We show the entire fine-tuning process (\emph{i.e.}, until overfitting). 
    ICO alleviates premature overfitting, resulting in a longer process.
    The model fine-tuned using manual prompts is prone to premature overfitting, and the generated images suffer from concept distortion. 
    ICO can mitigate these issues and achieve a better trade-off between concept fidelity and editability.
    We highlight the best generation with a yellow border for each query prompt. We can find that the optimal trade-off point varies for different query prompts, and the trade-off is hard to reconcile for customized generations with low model priors (\emph{e.g.}, sample (c)).
    DCI can effectively address these issues, generating satisfactory images for each prompt using one fine-tuned model.
    The prompt schemes of ICO used in “ICO” and “ICO+DCI” are different, which is detailed in Section~\ref{sec:exp_comp}.
    }
    \label{fig:compare}
\end{figure*}

For a specific model, its performance of customized generation is confined to a fidelity-editability trade-off curve and can only be adjusted along this curve by modifying some fine-tuning parameters (\emph{e.g.}, learning rate, fine-tuning step, \emph{etc.}).
Previous methods reluctantly seek a compromise between high concept fidelity and high editability, making it difficult to achieve both simultaneously.
Besides, the generative prior varies across different concepts and query prompts, leading to variations in the optimal trade-off point, as illustrated in Figure \ref{fig:compare}.
However, the raw fine-tuned models lack powerful inference-time adjustment options, making it challenging to achieve the best trade-off for all query prompts using a single fine-tuned model in practical applications.
Furthermore, as shown in~\Cref{fig:compare} (c), reconciling the trade-off becomes infeasible for customized generation with weak generative priors, such as challenging concepts (\emph{e.g.}, those with distributions far from the pre-trained domain) or rare prompted contexts.
To address these challenges, our goal is to liberate the fine-tuned model from the constraint of the fidelity-editability trade-off and achieve satisfactory generations for all query prompts using a single fine-tuned model, as shown in~\Cref{fig:trade_off} (b). 

In this paper, we propose a \emph{``\underline{D}ivide, \underline{C}onquer, then \underline{I}ntegrate''} (DCI) framework, which performs a surgical adjustment to overcome the fidelity-editability trade-off at inference.
In DCI, the two conflicting components in the trade-off are decoupled and individually conquered by two collaborative branches, namely the concept branch and the auxiliary branch.
Specifically, the concept branch is responsible for contributing the concept-relevant content with high fidelity, while the auxiliary branch is tasked with providing the concept-irrelevant content. 
Both branches contribute specialized contents from their entire generations, which are selectively integrated using a Dual-Branch Integration Module (DBIM) during the denoising steps to produce the final output with high concept fidelity and prompt alignment.
The integration ratio between the two branches is controllable, offering an inference-time adjustment  mechanism to seek the optimal trade-off for each query prompt.

The performance boundary of the proposed DCI is dependent on the overall performance of the raw fine-tuned model (\emph{i.e.}, the red line in~\Cref{fig:trade_off} (b)).
Previous works utilize manually crafted templates (\emph{e.g.}, {\menlo "A photo of \{\}"}) to construct text prompts for model fine-tuning.
We argue that manually designed prompts are fixed and may provide imprecise guidance for model adaptation, leading to a deterioration of the overall trade-off.
Therefore, we propose an \emph{\underline{I}mage-specific \underline{C}ontext \underline{O}ptimization} (ICO) fine-tuning strategy for model customization, which allows us to obtain a better fine-tuned model for DCI.
In ICO, we replace the manually designed prompt templates with learnable image-specific contexts for all reference images.
This prompt format is more effective at disentangling irrelevant content from the concept, adaptively providing a more precise fine-tuning direction to achieve a better overall performance, as illustrated by the comparison between ``Manual prompt'' and ``ICO'' in~\Cref{fig:compare}.
Extensive experiments demonstrate the effectiveness of DCI and ICO in reconciling the fidelity-editability trade-off, particularly for generations with weak generative priors.

\section{Related Work}
\myparagraph{Text-to-image Generation.} The purpose of text-to-image generation is to synthesize visually realistic images based on textual descriptions.
Early studies~\cite{ref9, ref10, ref11, ref12} translate natural language into images by employing generative adversarial networks~\cite{ref13}, which are constrained within the context of small-scale input scenarios.
Autoregressive models such as DALL-E~\cite{ref15}, Cogview~\cite{ref16}, NUWA~\cite{ref17}, and Parti~\cite{ref18}, reframe text-to-image generation as a sequence-to-sequence problem.
Recent advancements~\cite{ref19, ref20, ref4, ref21} in text-to-image generation have harnessed diffusion models (DMs)~\cite{ref22} as the generative backbone. These models generate images through a denoising task while incorporating text conditions during the denoising process.
Such successes lay the foundations for a multitude of text-guided image synthesis works~\cite{ref42, couairon2022diffedit, xue2023freestyle, zhang2023sine, huang2023collaborative, zhao2023magicfusion}.

\myparagraph{Prompt Tuning.}
Prompting is a parameter-efficient paradigm for adapting pre-trained models to a variety of downstream tasks~\cite{ref23}.
This line of research stems from the field of natural language processing (NLP). Previous studies in this area employ hand-crafted prompts~\cite{ref24, ref26} or generative prompts~\cite{ref25}, which can be inefficient and suboptimal.
To alleviate this problem, recent work has focused on prompt tuning (PT)~\cite{ref30, ref31, ref32}, 
which treats prompts as task-specific continuous vectors and optimizes them directly using gradients during the fine-tuning process.
With the remarkable advancement of pre-trained vision-language models~\cite{ref33, ref34} in recent years, prompt learning has also been introduced into various computer vision tasks~\cite{ref1, ref27, ref28, ref29, ref35, ref36, ref37, ref38, ref39, ref40, ref41, guo2023zero}.

\myparagraph{Customized Generation.}
Customized generation aims to incorporate novel concepts into pre-trained text-to-image models and generate images of these concepts in new contexts.
Pioneering works implant the concept by either inverting it into a word embedding vector~\cite{ref6} or bounding it with a special token~\cite{ref2}.
Subsequent works focus on developing more efficient methods~\cite{ref3, gal2023designing, han2023svdiff}, tuning-free methods~\cite{wei2023elite, xiao2023fastcomposer, chen2024subject}, for customized generation.
Some works focus on multi-concept customization~\cite{ref3, liu2023cones, han2023svdiff, gu2024mix}, which customize multiple new concepts and integrating them into a single image.
Some studies~\cite{zhou2023enhancing, hao2023vico} aim to enhance detail preservation in text-to-image customization.
Our work, instead, explores and addresses the fidelity-editability trade-off in customized generation.

\section{Method}\label{sec:method}

\subsection{Preliminary of Customized Generation}\label{sec:Preliminary}

The implantation of a custom concept $\mathcal{C}$ into pre-trained Diffusion Models (DMs) is accomplished through a text-to-image reconstruction task, using $N$ image-prompt pairs $\mathcal{D}=\left\{(\bm{X}_i, \bm{P}_i) \right\}_{i=1}^N $. 
$\bm{X}_i$ and $\bm{P}_i$ are reference images and the corresponding textual prompts of $\mathcal{C}$.
We utilize Latent Diffusion Models (LDMs)~\cite{ref4} as the generative backbone.
Prompt $\bm{P}_i$ is firstly projected to an intermediate representation $\mathbf{c}=\tau_\theta({\bm{P}_i})\in \mathbb{R}^{M\times d_c}$ by a text encoder $\tau_\theta$. 
$\mathbf{c}$ is then injected into the LDMs via the cross-attention mechanism~\cite{ref7}, serving as the conditioning for the reconstruction of image $\bm{X}_i$. The reconstruction process is regularized by the following squared error loss:
\begin{equation}\label{eq:ori_obj}
    \mathbb{E}_{\mathbf{z}, \bm{\epsilon} \sim \mathcal{N}(0,1),\mathbf{c}, t}\left[w_t \Vert \bm{\epsilon} - \bm{\epsilon}_\theta (\mathbf{z}_t, \mathbf{c}, t) \Vert_2^2  \right]  \,,
\end{equation}
where $\mathbf{z}_t := \alpha_t \mathbf{z} + \sigma_t \bm{\epsilon}$ is the noised latent code at timestep $t$, $\mathbf{z}$ is the clean latent code of the training data, and $w_t$, $\alpha_t$, $\sigma_t$ are terms that determine the loss weight and noise schedule;
$\bm{\epsilon}_\theta$ is a denoising autoencoder implemented using a conditional U-Net~\cite{ref5}.

During inference, given the query text $\bm{P}_q(\mathcal{C})$ containing the learned concept $\mathcal{C}$, the customized DMs can generate the corresponding images conditioned on $\mathbf{c} = \tau_\theta(\bm{P}_q(\mathcal{C}))$. 
Concretely, an initial noise map $\mathbf{z}_T\sim\mathcal{N}(0,1)$ is iteratively denoised from timestep $t=T$ to $t=0$ in the latent space using either deterministic samplers~\cite{ref43} or stochastic sampler~\cite{ho2020denoising}.
Then $\mathbf{z}_0$ is decoded to image space using an decoder $f_D$ to generate the target image $\hat{\mathbf{x}}= f_D(\mathbf{z}_0)$.

\subsection{Learning to Prompt for Customized Fine-tuning}\label{sec:learntoprompt}

In previous works~\cite{ref2, ref3, ref6}, the textual prompts $\bm{P}$ used in model fine-tuning are constructed from manually crafted templates (\emph{e.g.}, {\menlo "A photo of \{\}"}).
We argue that the manually designed part in prompts may provide imprecise guidance for model adaptation, resulting in a deterioration of the overall performance.

\begin{wrapfigure}{r}{7cm}
    \centering
    \includegraphics[width=0.95\linewidth]{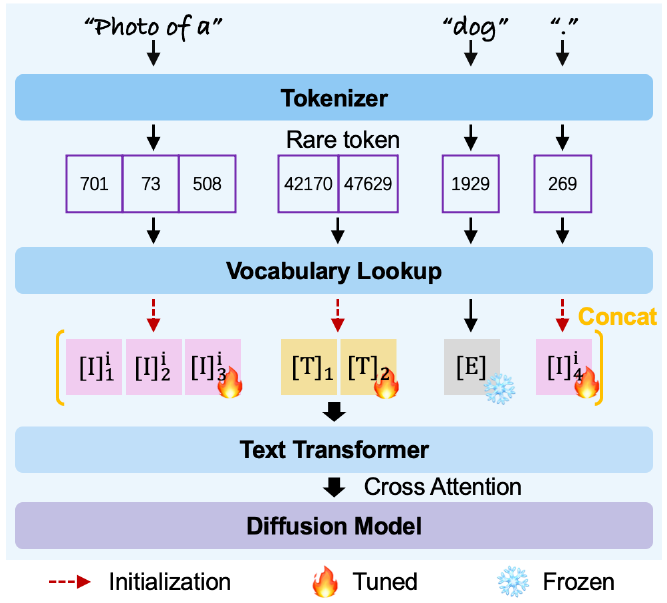}
    \caption{\textbf{Overall framework of the prompt generating process in fine-tuning.}
    }
    \label{fig:fig1}
\end{wrapfigure}

We redesign the prompts of the reference images, and an example of the prompt generating process is illustrated in~\Cref{fig:fig1}.
We introduce learnable image-specific contexts to the prompt of each reference image, completely replacing the manually designed templates that remain fixed during the fine-tuning.
We refer to this fine-tuning strategy as \emph{\underline{I}mage-specific \underline{C}ontext \underline{O}ptimization} (ICO).
Specifically,  the prompts input to the text transformer are designed as follows:
\begin{equation}
    \label{single_prompt}
    \bm{P}_i = \left[ \mathbf{I}\right]_1^i  
    \cdots \left[ \mathbf{I}\right]_M^i \left[ \mathbf{T}\right]_1 \cdots \left[ \mathbf{T}\right]_S \left[ \mathbf{E} \right] \left[ \mathbf{I}\right]_{M+1}^i  
    \cdots \left[ \mathbf{I}\right]_{M+L}^i\,,
\end{equation}
where $\bm{P}_i$ is the prompt of image $\bm{X}_i$ and consists of three different components: image-specific context vectors $\left[ \mathbf{I}\right]_m^i (m=1, \ldots, M+L) \in \mathbb{R}^d $, concept-specific context vectors $\left[ \mathbf{T}\right]_s (s=1, \ldots, S) \in \mathbb{R}^d$, and the concept descriptor $\left[ \mathbf{E} \right] \in \mathbb{R}^{t \times d}$. 
The learned concept-specific context $\mathbf{T}$, in combination with the concept descriptor, is utilized to generate images of new concept at inference.

\myparagraph{Components of Fine-tuning Prompt.} \emph{Image-specific Contexts} $\left[ \mathbf{I}\right] $ are continuous vectors and are independent for each image. 
These vectors can be optimized end-to-end during fine-tuning, allowing them to model the image-specific content adaptively and offer more precise guidance for model adaptation.
These context vectors can be positioned either at the beginning or the end of the entire prompt, serving different purposes in the learning process~\cite{ref1}. 
\emph{Concept-specific Context} $\left[ \mathbf{T}\right]$, also represented as {\menlo "V$*$"}, shares a learnable congruous context with all reference images, which is used to model the new concept~\cite{ref6, ref3}.
\emph{Concept Descriptor} $\left[ \mathbf{E} \right]$ is the word embedding of a coarse class noun {\menlo "<noun>"} that describes the new concept~\cite{ref2}, \emph{e.g.}, embedding of {\menlo "cat"}.
The design scheme of the fine-tuning prompt possesses flexibility in the position and length of image-specific and concept-specific contexts. 

\subsection{Trade-off Reconciled at Inference}

\begin{figure*}[!t]
    \centering
    \includegraphics[width=\linewidth]{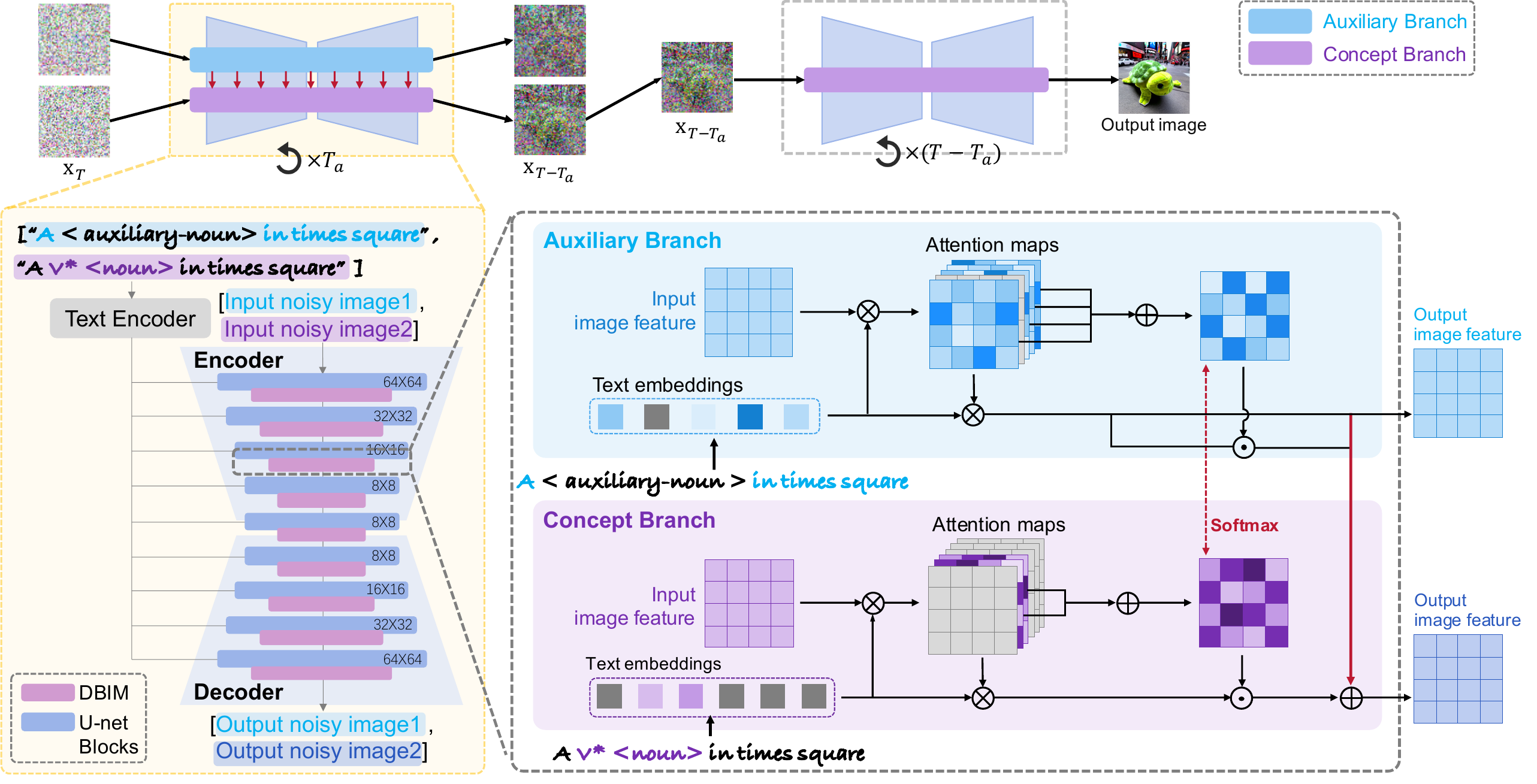}
    \caption{\textbf{An overview of the proposed \emph{``Divide, Conquer, then Integrate''} (DCI) framework.} DCI divides the customized generation task into two collaborative branches, namely the concept branch and the auxiliary branch. The concept branch takes the original query prompt as its conditioning and responsible for generating concept with high fidelity. The auxiliary branch takes an auxiliary prompt as its conditioning and contributes concept-irrelevant content. During the latent denoising steps, the contents generated by two branches are selectively integrated using a Dual-Branch Integration Module (DBIM) in each cross-attention layer.
    }
    \label{fig:fig4}
\end{figure*}

While the prompt tuning strategy improves the overall performance of the fine-tuned model, the fidelity-editability trade-off still persists.
This issue becomes infeasible to reconcile for customized generation with weak model priors, such as challenging concepts (\emph{e.g.}, those with distributions far from the pre-trained domain) or rare prompted contexts.
We propose a novel framework called \emph{``\underline{D}ivide, \underline{C}onquer, then \underline{I}ntegrate''} (DCI), which performs latent feature rectification to liberate the fine-tuned model from the trade-off at inference.

In the early stage of denoising ($T\sim(T-T_a)$), DCI \emph{divides} the generation task and \emph{conquers} it with two collaborative branches, namely the concept branch $\mathcal{B}_c$ and the auxiliary branch $\mathcal{B}_a$. $\mathcal{B}_c$ and $\mathcal{B}_a$ share the same model parameters.
The overview of DCI is shown in~\reffig{fig4}.
Concretely, for a given prompt $\mathcal{P}$ of a custom concept {\menlo V$*$ <noun>} (\emph{e.g.}, $\mathcal{P}$ is {\menlo "A V$*$ <noun> in times square"}), we first construct an auxiliary prompt $\mathcal{P}_a$ by replacing {\menlo "V$*$ <noun>"} within $\mathcal{P}$ with an auxiliary class noun {\menlo "<auxiliary-noun>"}, while leaving the rest unaltered (\emph{i.e.}, $\mathcal{P}_a$ is {\menlo "A <auxiliary-noun> in times square"}).
$\mathcal{P}$ and $\mathcal{P}_a$ are then taken as the text conditions for the concept branch $\mathcal{B}_c$ and the auxiliary branch $\mathcal{B}_a$.
The auxiliary branch can regain powerful editability with the assistance of auxiliary nouns (\emph{cf.} \textbf{Auxiliary Noun.}).
The fine-tuned model used in DCI is solely required to excel at concept modeling, enabling the concept branch to generate concept-relevant content with high visual fidelity.
These two branches are simultaneously forwarded within the diffusion model.
Both branches contribute specialized components from their complete generations, with the concept branch providing concept-relevant content and the auxiliary branch offering concept-irrelevant content.
These two branches are selectively \emph{integrated} in the early stage of denoising using a \emph{\underline{D}ual-\underline{B}ranch \underline{I}ntegration \underline{M}odule} (DBIM). 
We set $T_a = 0.15T$ and provide a quantitative analysis of the effect of $T_a$ in the supplementary material.

\myparagraph{Auxiliary Noun.}
A key observation is that the editability reduction during model customization originates from the tokens associated with the custom concept, including the entire {\menlo "V$*$ <noun>"} and the noun {\menlo "<noun>"}, rather than from the fine-tuned model itself, as discussed in supplementary material.
Therefore, {\menlo "<auxiliary-noun>"} need to be a noun distinct from {\menlo "<noun>"}.
Replacing {\menlo "V$*$ <noun>"} with {\menlo "<auxiliary-noun>"} empowers the fine-tuned model to regain the powerful editability over the query contexts, thus the content synthesized by the auxiliary branch maintains high prompt alignment.
Additionally, we can select nouns that have a high probability of occurrence within the context of the query prompt to further enhance textual editability, as explored in Section~\ref{sec:brk_limi}.

\myparagraph{Dual-Branch Integration Module.}
To selectively integrate the generated content that each branch is responsible for, we introduce a \emph{Dual-Branch Integration Module} (DBIM) to each cross-attention layer within the U-Net, as depicted in \reffig{fig4}. We elaborate on the design of DBIM in the following content.

We first introduce the cross-attention layer in LDMs briefly. The cross-attention layer utilizes latent image features $\mathcal{F}$ and text embeddings $\mathcal{E}$ to compute queries $Q=\ell_Q (\mathcal{F})$, keys $K=\ell_K (\mathcal{E})$, and values $V=\ell_V (\mathcal{E})$ through three projection layers: $\ell_Q$, $\ell_K$ and $\ell_V$. Then the spatial attention maps $\mathcal{M} \in \mathbb{R}^{C \times H \times W}$ are calculated as $\mathcal{M} = QK^T/ \sqrt{d}$,
where $d$ is the dimension of the queries and keys, while $C$, $H$, $W$ are the channel dimension, height, and weight of $\mathcal{M}$, respectively. Finally, the output of the cross-attention layer is obtained as $\mathcal{\hat{\mathcal{F}}} = \mathcal{M}V$. 

It is worth noting that each textual token in the conditional prompt is associated with a specific channel of the attention map $\mathcal{M}$, which determines the spatial layout and geometry of the corresponding textual semantics in the generated image~\cite{ref42, zhao2023magicfusion}.
Inspired by this, we construct token sets containing the content that each branch is responsible for.
The token set for $\mathcal{P}$ is {\menlo $\mathcal{T}(\mathcal{P})=$ $\{$"V$*$"$, $"<noun>"$\}$}, and the token set for $\mathcal{P}_a$ consists of all tokens in $\mathcal{P}_a$ except for {\menlo "<auxiliary-noun>"}, denoted as {\menlo $\mathcal{T}(\mathcal{P}_a)= \mathbb{U}_{\mathcal{P}_a} \backslash \{$"<auxiliary-noun>"$\}$}.
In the $i$-th cross-attention layer during the denoising step $t$,
we use the attention maps associated with the tokens in the token sets $\mathcal{T}(\mathcal{P})$ and $\mathcal{T}(\mathcal{P}_a)$ to infer the dynamic aggregation mask $\mathrm{M}_{\mathcal{B}_c}^{t,i}$ and $\mathrm{M}_{\mathcal{B}_a}^{t,i}$ for branch $\mathcal{B}_c$ and $\mathcal{B}_a$ :
\begin{equation}
    \mathrm{M}_{\mathcal{B}_c}^{t,i} = \sum_{k \in \mathcal{T}(\mathcal{P})} \mathcal{M}_k^{t,i}
    \,, \mathrm{M}_{\mathcal{B}_a}^{t,i} = \sum_{k \in \mathcal{T}(\mathcal{P}_a)} \mathcal{M}_k^{t,i} \,,
\end{equation}
where $\mathcal{M}_k^{t,i}$ is the attention map related to token $k$.

Subsequently, we first perform softmax $\mathbf{S}_{s}$ in the spatial dimension to sharpen the spatial distribution of each mask, and then apply softmax $\mathbf{S}_{c}$ across two branches at each pixel to regularize the overall aggregation strength.
The final aggregation masks $\hat{\mathrm{M}}^{t,i}$ are defined as follows:
\begin{equation}
    \label{eq:mask_calcute}
     \left( \hat{\mathrm{M}}_{\mathcal{B}_c}^{t,i}, \hat{\mathrm{M}}_{\mathcal{B}_a}^{t,i}  \right) =
    \mathbf{S}_{c} \left(  \mathbf{S}_{s} \left( \mathrm{M}_{\mathcal{B}_c}^{t,i} \right), \lambda \mathbf{S}_{s} \left( \mathrm{M}_{\mathcal{B}_a}^{t,i} \right)      \right) \,,
\end{equation}
where $\lambda$ is a factor that can control the injection strength of concept-irrelevant information at inference time.

$\hat{\mathrm{M}}^{t,i}$ dynamically determines the spatially varying influences of each generative branch, allowing for the adaptive suppression or enhancement of the relative expression strength at different spatial regions.
The output of DBIM is a linear combination of the two generative branches that weighted by $\hat{M}^{t,i}$: 
\begin{equation}
    \label{eq:branch_agg}
    \hat{\mathcal{F}}^{t,i} = \hat{\mathcal{F}}_{\mathcal{B}_c}^{t,i} \odot \hat{\mathrm{M}}_{\mathcal{B}_c}^{t,i} + \hat{\mathcal{F}}_{\mathcal{B}_a}^{t,i} \odot \hat{\mathrm{M}}_{\mathcal{B}_a}^{t,i}
     \,,
\end{equation}
where $\hat{\mathcal{F}}_{\mathcal{B}_c}^{t,i}$ and $\hat{\mathcal{F}}_{\mathcal{B}_a}^{t,i}$ are the raw outputs of the concept branch and auxiliary branch at the $i$-th cross-attention layer in denoising step $t$, and $\odot$ denotes pixel-wise multiplication. $\hat{\mathcal{F}}^{t,i}$ is used to update the concept branch output $\hat{\mathcal{F}}_{\mathcal{B}_c}^{t,i}$.

\section{Experiments}\lblsec{expr}

\subsection{Experimental Setup}
\myparagraph{Datasets.}
We conduct experiments on $16$ datasets from previous
works~\cite{ref6, ref2, ref3}, each of which contains several reference images of a unique subject. These subjects cover a diverse range of categories, including pets, toys, plushies, \emph{etc}.
Each subject is evaluated using different types of query prompts, such as recontextualization, artistic variation, and attribute modification.

\myparagraph{Evaluation Metrics.} Concept reconstruction capacity and editability are the two primary metrics for evaluating customized generation~\cite{ref6, ref2, ref3}.
The visual alignment of the generated images with the target concept reflects the model's reconstruction capacity, which is quantified using the average pairwise cosine similarity between the CLIP~\cite{ref33} embeddings of the generated and reference images.
Editability is represented by the text alignment of the generated images with the query prompts.
We measure this metric using the average CLIP-space cosine similarity between the generated images and their textual prompts, with the exclusion of the task-specific context {\menlo "V$*$"}.
We also calculate the DINO-based image similarity~\cite{ref2} and present it in the supplementary material.

\myparagraph{Baselines.}
We compare our method with Textual Inversion~\cite{ref6}, DreamBooth~\cite{ref2}, and Custom Diffusion~\cite{ref3}.
Textual Inversion~\cite{ref6} inverts the new concept into a special token.
DreamBooth~\cite{ref2} fine-tunes the entire diffusion model to bind the new concept with a fixed identifier.
Custom Diffusion~\cite{ref3} introduces a new token for each novel concept and jointly optimizes the token embedding along with a subset of parameters in cross-attention layers.
The implementation details of our method and baselines are provided in the supplementary material.

\subsection{Comparisons}\label{sec:exp_comp}

\begin{figure*}[!t]
    \centering
    \includegraphics[width=\linewidth]{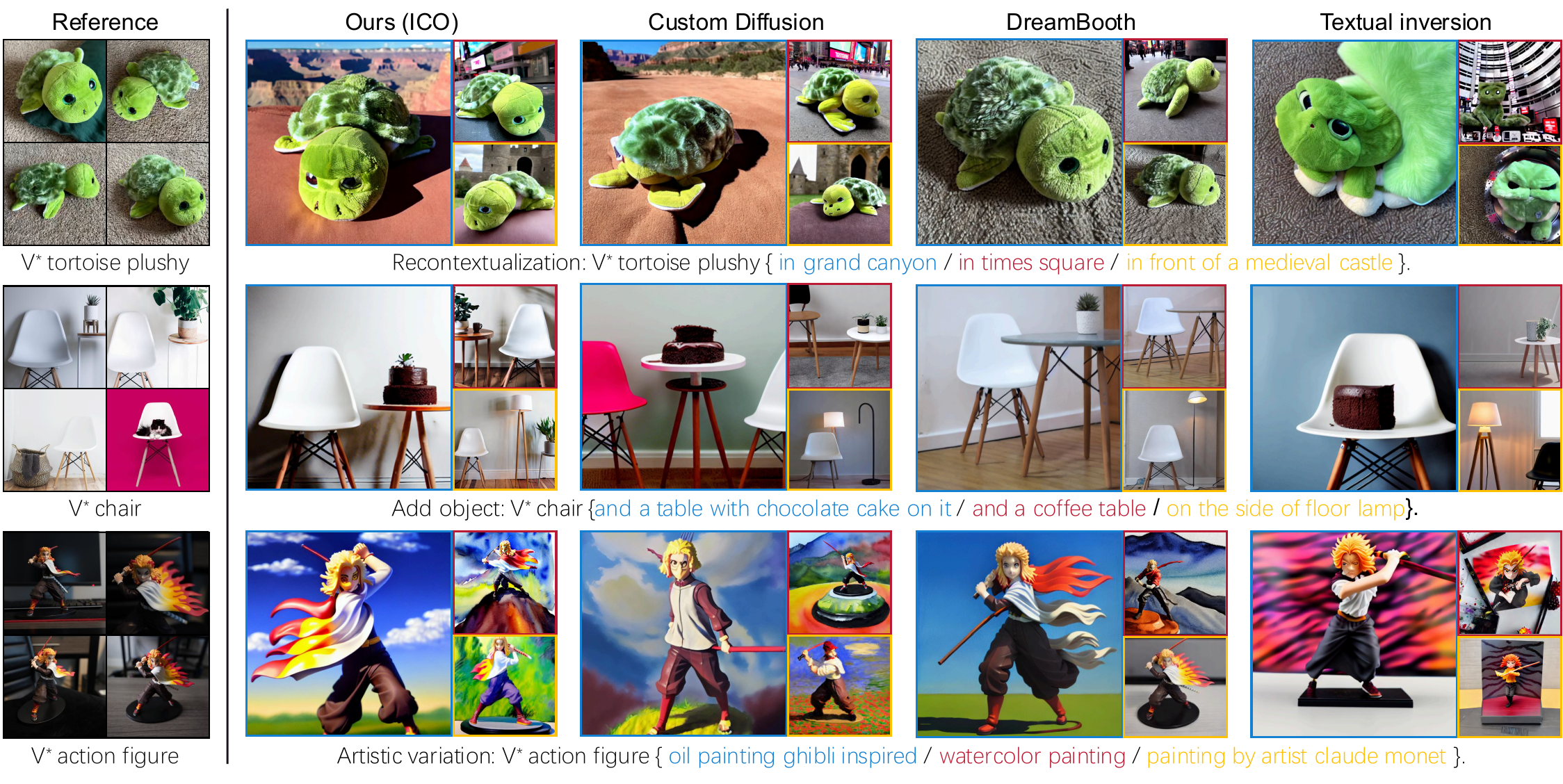}
    \caption{{\textbf{Visual comparsion of ICO with baselines.} 
    We illustrate the effectiveness of ICO by comparing it to three baselines. Reference images are shown on the left.
    \textit{ First row}: customizing various scenes for the target concept. Our method can better reconstruct the target concept while maintaining high text alignment.
    \textit{Second row} : adding new objects. \textit{Third row} : customizing artistic styles for the target concept. While other methods suffer from concept distortion, our method can generate the target concept with higher concept fidelity.
    } }
    \label{fig:results_single}
\end{figure*}

\begin{figure}[H]
    \centering
    \includegraphics[width=\linewidth]{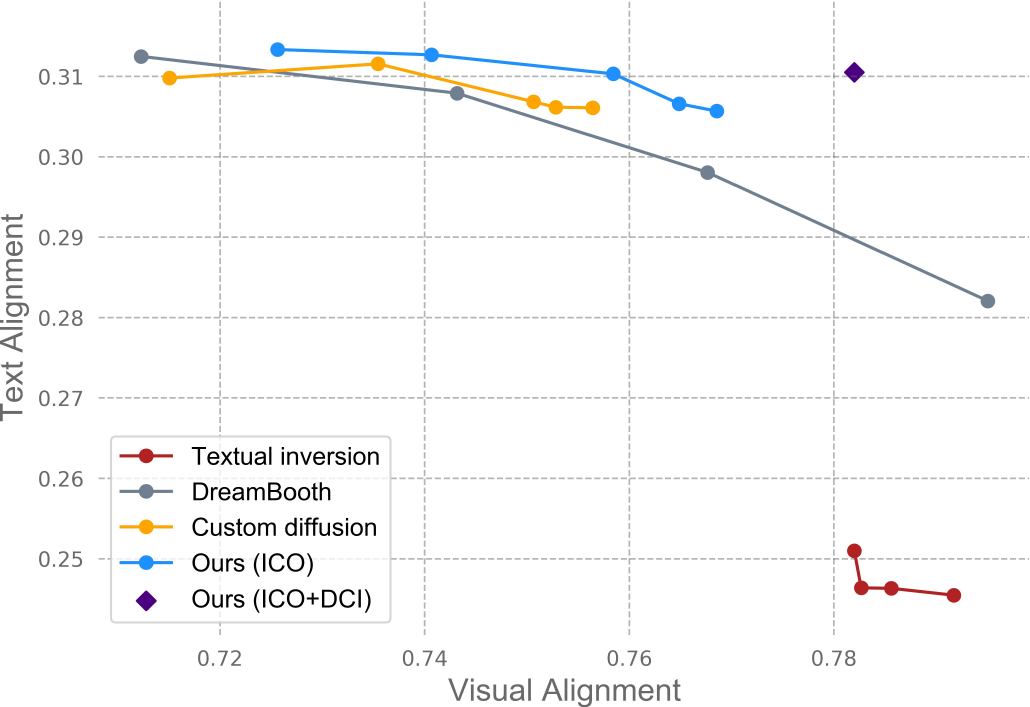}
    \caption{\textbf{Quantitative comparisons.} The quantitative evaluation results of our method, Textual Inversion, DreamBooth, and Custom Diffusion. We mark the different evaluation points.}
    \label{fig:fig6}
\end{figure}

\myparagraph{Comparisons of ICO with Baselines.}
We evaluate the proposed ICO in this section.
We utilize the following prompt scheme to strike a balance between concept fidelity and editability: $M=4$, $L=2$ and $S=2$ (cf. ablation study). 
Since it is challenging to determine the optimal trade-off point during the fine-tuning process for different methods, we evaluate all methods at various fine-tuning steps (until convergence or overfitting) for a more comprehensive comparison.
The quantitative evaluation results are depicted in~\reffig{fig6}. We mark the different evaluation points.
We can observe that the fine-tuning process of each method delineates a fidelity-editability trade-off curve, and ICO achieves the best performance in balancing concept reconstruction and prompt adherence.
We present sample generations of ICO and the other three baselines in~\Cref{fig:results_single}. Comparing to baselines, ICO better captures both the high-level semantics and the visual details of the target concept, while also demonstrating higher text-alignment for various types of prompts.

\myparagraph{Comparisons of DCI with Baselines.}
In this section, we evaluate the effectiveness of DCI.
DCI is an inference-time framework that requires combining with a specific fine-tuning strategy, which in our experiment is ICO.
To empower the concept branch with high reconstruction capacity, the models used for DCI 

\begin{multicols}{2}

\begin{figure*}[!t]
    \centering
    \includegraphics[width=\linewidth]{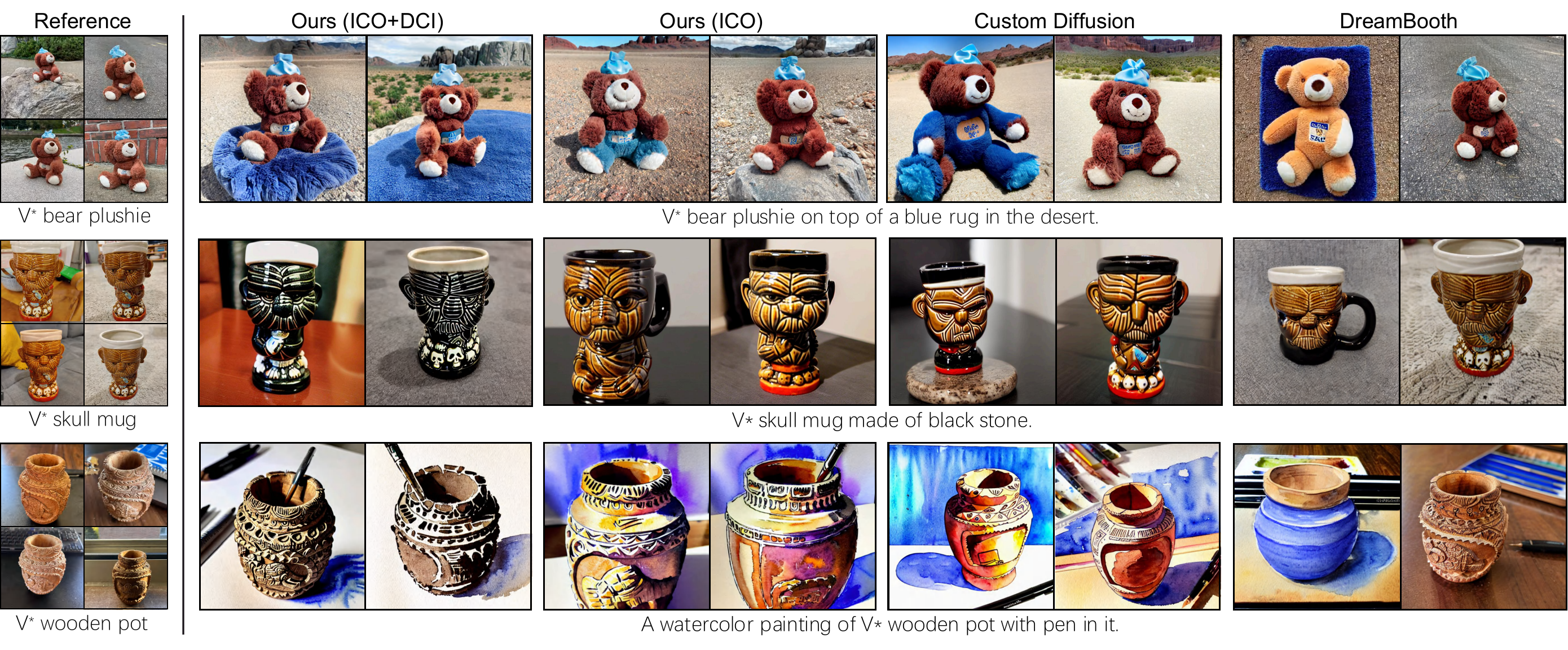}
    \caption{{\textbf{Visual comparisons of DCI with baselines.}}
    Reference images are shown on the left. We show sample generations for three types of query prompts: Recontextualization (\textit{first row}), Property modification (\textit{second row}), and Artistic variations \& Add object (\textit{third row}).
    In the context of customized generation with weak model priors, the outputs of the raw fine-tuned models either suffer from severe subject distortions or overfit to the given subject.
    DCI can significantly alleviate this fidelity-editability trade-off. The images generated by DCI can preserve subject details while adhering faithfully to the query prompt.
    }
    \label{fig:fig9}
\end{figure*}

\begin{table}[H]
\centering
\setlength{\tabcolsep}{5pt}
\resizebox{\linewidth}{!}{
\begin{tabular}{lcccc}
\toprule
 & \thead{\textbf{Metric}} & \makecell{\textbf{Textual} \\ \textbf{Inversion}} & \thead{\textbf{DreamBooth}} & \makecell{\textbf{Custom} \\ \textbf{Diffusion}} \\
\midrule
\multirow{4}{*}{\shortstack[c]{\textbf{ICO}} }  & \multicolumn{1}{@{} l}{ \multirow{2}{*}{\shortstack[c]{Text\\ Alignment }} }    & \multirow{2}{*}{\shortstack[c]{ \textbf{84.38} $\%$ }}  & \multirow{2}{*}{\shortstack[c]{ \textbf{58.33} $\%$ }}  &  \multirow{2}{*}{\shortstack[c]{ \textbf{55.21} $\%$ }}   \bigstrut[t]\\
& &   & &  \\
\cdashline{2-5}
& \multicolumn{1}{@{} l}{ \multirow{2}{*}{\shortstack[c]{Visual\\ Alignment }} }   & \multirow{2}{*}{\shortstack[c]{ \textbf{71.88} $\%$ }}  & \multirow{2}{*}{\shortstack[c]{ \textbf{53.13} $\%$ }}  & \multirow{2}{*}{\shortstack[c]{ \textbf{57.29} $\%$ }}  \bigstrut[t] \\
&  &  &  &  \\
\midrule
\multirow{4}{*}{\shortstack[c]{\textbf{ICO}\\ \textbf{+DCI} } } 
&   \multicolumn{1}{@{} l}{ \multirow{2}{*}{\shortstack[c]{Text\\ Alignment }} }   & \multirow{2}{*}{\shortstack[c]{ \textbf{89.58} $\%$ }}  & \multirow{2}{*}{\shortstack[c]{ \textbf{66.67} $\%$ }}  & \multirow{2}{*}{\shortstack[c]{ \textbf{63.54} $\%$ }}  \bigstrut[t] \\
& &  &  &  \\
\cdashline{2-5}
& \multicolumn{1}{@{} l}{ \multirow{2}{*}{\shortstack[c]{Visual\\ Alignment }} } & \multirow{2}{*}{\shortstack[c]{ \textbf{76.04} $\%$ }}  & \multirow{2}{*}{\shortstack[c]{ \textbf{60.42} $\%$ }}  & \multirow{2}{*}{\shortstack[c]{ \textbf{68.75} $\%$ }}  \bigstrut[t]\\
& & & &  \\
\bottomrule
\end{tabular}
}
\caption{\textbf{User study.} 
In each paired comparison, our method is preferred ($\geq 50\%$) over the baseline methods in terms of both visual alignment and text alignment. The combination of DCI and ICO demonstrates a overwhelming user preference over other methods.}
\label{tbl:user_study}
\end{table}
\end{multicols}

\noindent are fine-tuned using the following prompt scheme: $M=1$, $ L=1$, and $S=3$ (cf. ablation study). 
Since DCI incurs additional denoising steps $T_a$ at inference, we reduce the total denoising steps (\emph{i.e.}, $T_a+T$) of DCI to match other single-branch baselines' steps for fair comparison.
We depicted the quantitative evaluation results of DCI in~\reffig{fig6}, comparing with the other three baselines and the raw model fine-tuned using ICO.
We can observe that DCI overcomes the inherent fidelity-editability trade-off in the fine-tuned model at inference time. It achieves both higher concept fidelity and text alignment compared to the other raw fine-tuned models.
We show sample generations of our method and the other baselines in~\reffig{fig9}.
The depicted generations of baselines are selected from all evaluation steps.
As for DCI, images are generated from one fine-tuned model for each concept.
As we can see, the raw fine-tuned models struggle to reconcile the trade-off, particularly for generations with weak model priors. These models either suffer from severe concept distortions or overfit to the given concept throughout the entire customization process.
DCI can overcome these inherent limitations of fine-tuned models and generate satisfactory images.
The target concepts are successfully rendered in the query contexts with high visual fidelity by DCI, while still maintaining a high variability and diversity in poses and articulations.
We also compare our method to ED-LoRA~\cite{gu2024mix} in the supplementary material.

\myparagraph{User Study.}
We further evaluate the proposed method through a user study. We conduct paired tests comparing the two versions of our method, namely ``ICO'' and ``ICO+DCI'', with DreamBooth, Textual Inversion, and Custom Diffusion.
In each comparison, users are presented with reference images $\mathcal{I}$, the query prompt $\mathcal{P}$, and two corresponding generations from two compared method (ours and baseline).
Two questions are designed to compare the visual alignment and text alignment of the two images.
For the visual alignment comparison, users are asked to select the better image by answering the question-``which image contains the subject that is more similar to the provided reference images.''. In the text alignment comparison, users are asked to select the better image with the question-``which image is more consistent with the textual description of the query prompt.''.
A total of 576 comparison results are collected for the evaluation, and the aggregated result is shown in~\reftbl{user_study}. 
As observed, our method, particularly the ``ICO+DCI'' version, demonstrates a dominant user preference over other methods.
This further validates the effectiveness of the proposed method in enhancing the overall performance of customized generation.

\subsection{Ablation Studies}\label{sec:ablation}
\myparagraph{Prompt Schemes.} 
We explore the impact of prompt design schemes on concept customization using a subset of subjects, focusing on three aspects: suffix length $M$, prefix length $L$, and concept-specific context length $S$.
We fine-tune Stable Diffusion~\cite{ref4} using different prompt schemes, comparing them to baselines that

\begin{wrapfigure}{r}{6cm}
    \centering
    \includegraphics[width=\linewidth]{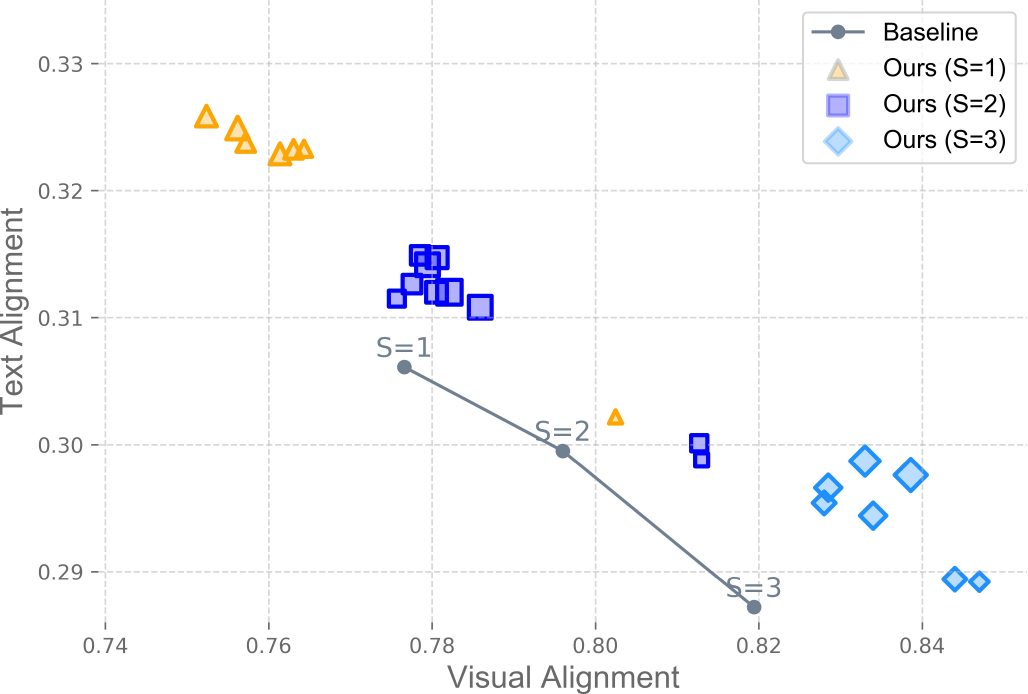}
    \caption{{\textbf{Investigations of ICO's prompt scheme.} $S$ is the length of concept-specific contexts. A larger data point size indicates a larger $M+L$ (the length of image-specific contexts).
    } }
    \label{fig:fig5}
\end{wrapfigure}

\noindent are fine-tuned using manually fixed prompt templates with $S=1$, $S=2$, and $S=3$.
As shown in~\reffig{fig5}, our method and baselines outline fidelity-editability trade-off curves, with our method achieving a better overall performance. This validates our hypothesis that replacing manual prompt templates with learnable contexts can improve model customization.
Besides, the prompt scheme impacts model convergence.
In general, $M+L$ (the total length of image-specific contexts) influences the overall trade-off (distance to the origin).
$S$ affects model's convergence point on trade-off curve.
An increase in $S$ enhances the model's concept reconstruction capacity but diminishes its editability. 

\begin{figure}[H]
    \centering
    \includegraphics[width=\linewidth]{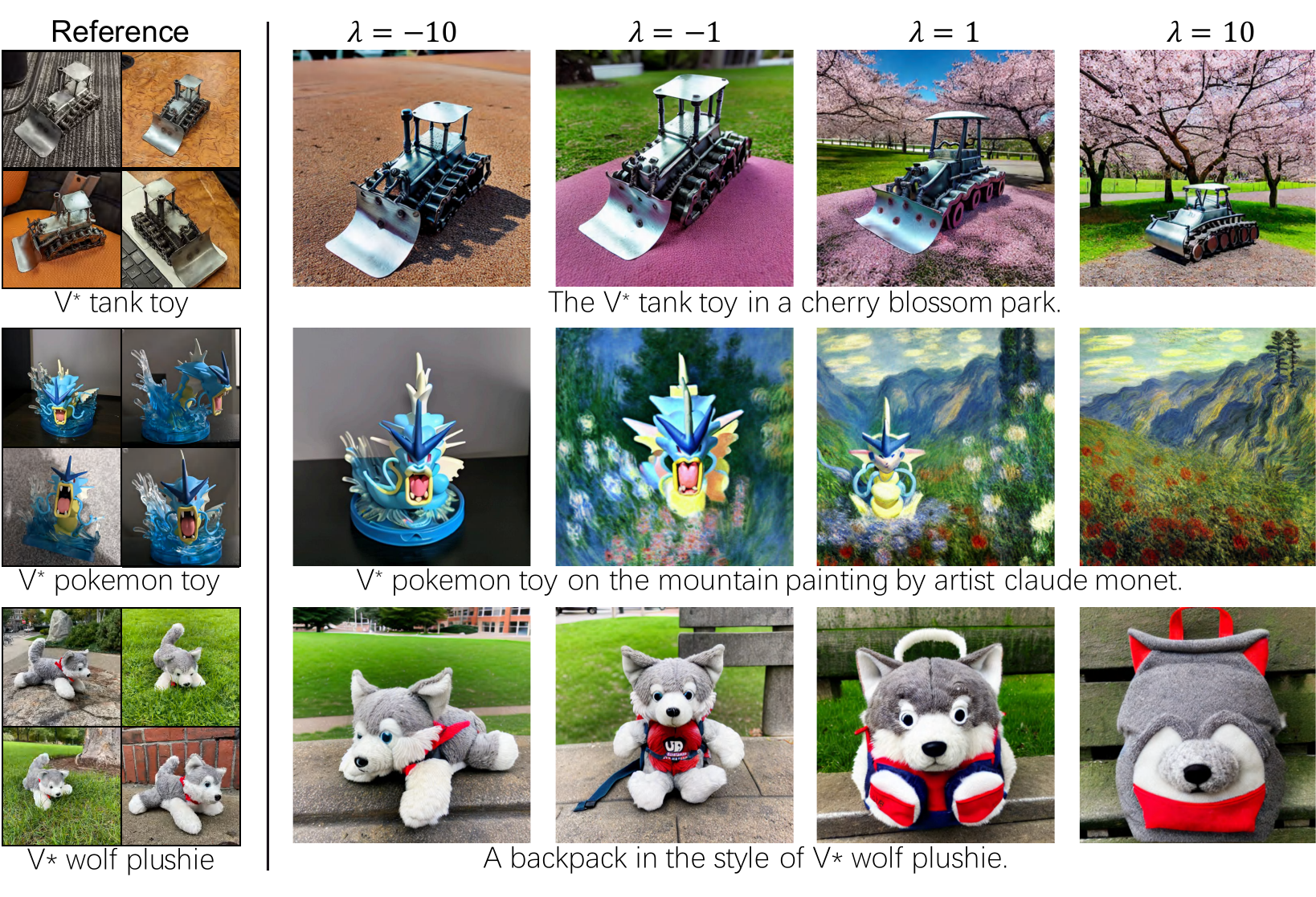}
    \caption{{\textbf{Comparisons of generations with different injection strength $\lambda$.}  The reference images are shown on the left. The values of $\lambda$ used in DCI are indicated at the top of each column. 
    } }
    \label{fig:fig7}

\end{figure}

\myparagraph{Injection Strength $\bm{\lambda}$.}
In DCI, the injection ratio ${\lambda}$ in Eqn.~\ref{eq:mask_calcute} controls the injection strength of the concept-irrelevant content, which
offers users with flexibility for inference-time adjustment.
We explore the influence of different values of $\lambda$ on customized generation and provide visual examples in~\reffig{fig7}.
Specifically, the sample generations closely resemble reference images when $\lambda$ is small. As $\lambda$ gradually increases, more concept-irrelevant content is injected into the concept branch, leading to a significant improvement in text alignment and a slight decrease in concept fidelity.
When $\lambda$ becomes too large, concept fidelity will sharply decline. This inference-time adjustment mechanism allows users to obtain the optimal generation for each individual query prompt.
In practical applications, $\lambda$ is typically set to $-2\sim2$ based on user demand.

\myparagraph{Ablation of Key Components in DCI.} We ablate the following three components in DCI: 1) We use masks in Eqn.~\ref{eq:branch_agg} to selectively aggregate two branches.

\begin{multicols}{2}

\begin{figure}[H]
    \centering
    \includegraphics[width=\linewidth]{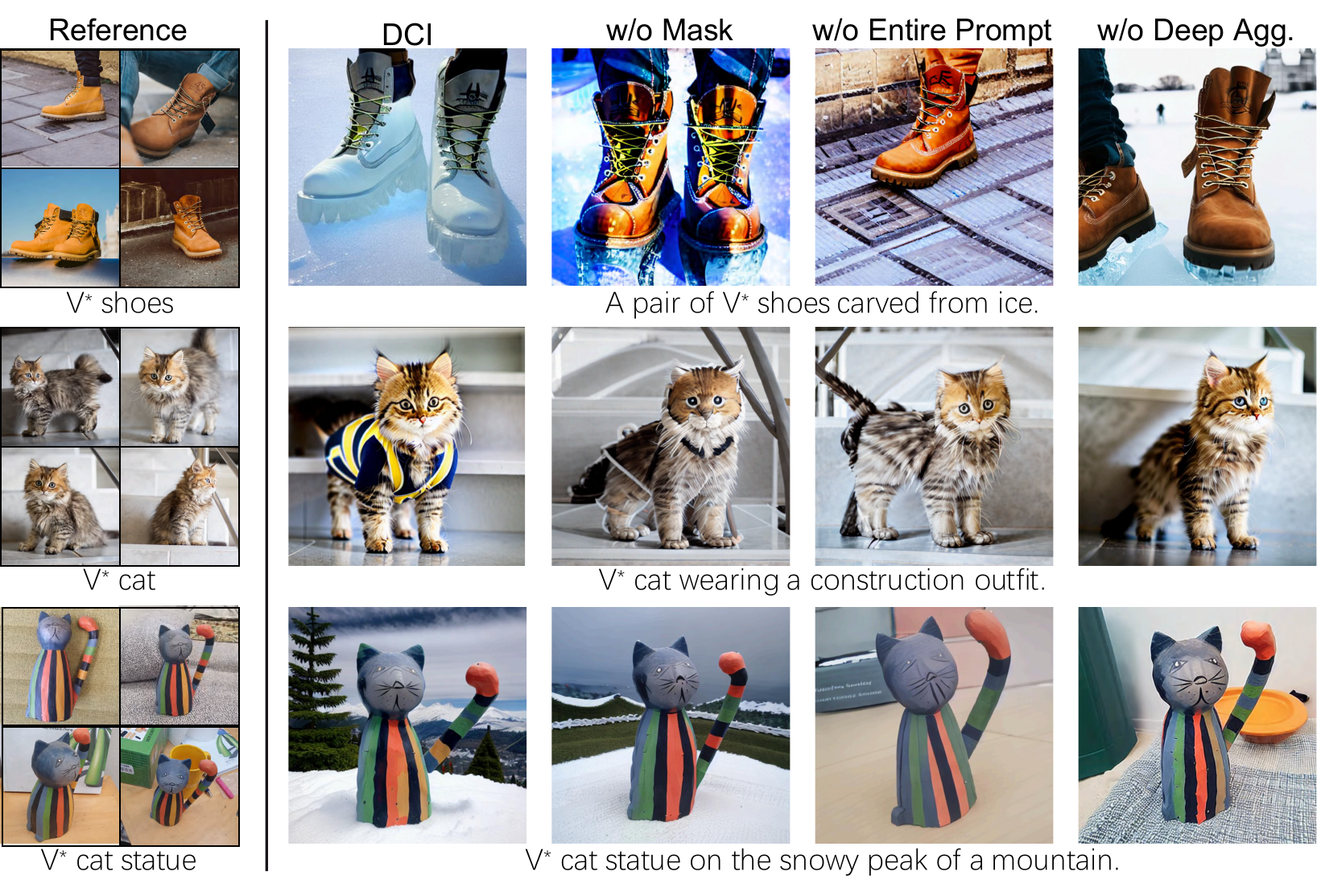}
    \caption{{\textbf{Visual comparisons of different variants.}  The reference images are shown on the left. 
    Each column verifies the necessity of one component, which is indicated at the top.
    } }
    \label{fig:abla_comp}

\end{figure}

\end{multicols}

\noindent
In the \emph{``w/o mask''} scenario, we verify the necessity of masks by ablating it and directly adding two branches.
2) We use entire prompts as the conditioning for each branch in DCI. In the \emph{``w/o Entire Prompt''} scenario, we generate samples by simply using the content that each branch is responsible for as its prompt. For example, for the query prompt {\menlo "V* shoes carved from ice"}, we use {\menlo "V* shoes"} and {\menlo "carved from ice"} as the prompts for the concept branch and the auxiliary branch.
3) We aggregate two branches in each cross-attention layer within the U-Net. In the \emph{``w/o Deep Agg.''} scenario, we aggregate two branches by only adding the output of the last U-Net cross-attention layer.
We show sample generations in~\reffig{abla_comp}.
As we can see, without masks, the influence scope of each branch extends to the entire image, resulting in integration confusion and subject distortion. The use of the entire prompt is essential for determining and aligning the layout of the images in two branches. When the prompt is separated, the two branches fail to target the correct regions they are responsible for and may generate artifacts.
Finally, the injection capacity of the auxiliary branch will significantly degrade without deep integration.

\subsection{Visualization of Aggregation Masks}\label{sec:vis_mask}
\begin{figure}[H]
    \centering
    \includegraphics[width=0.99\linewidth]{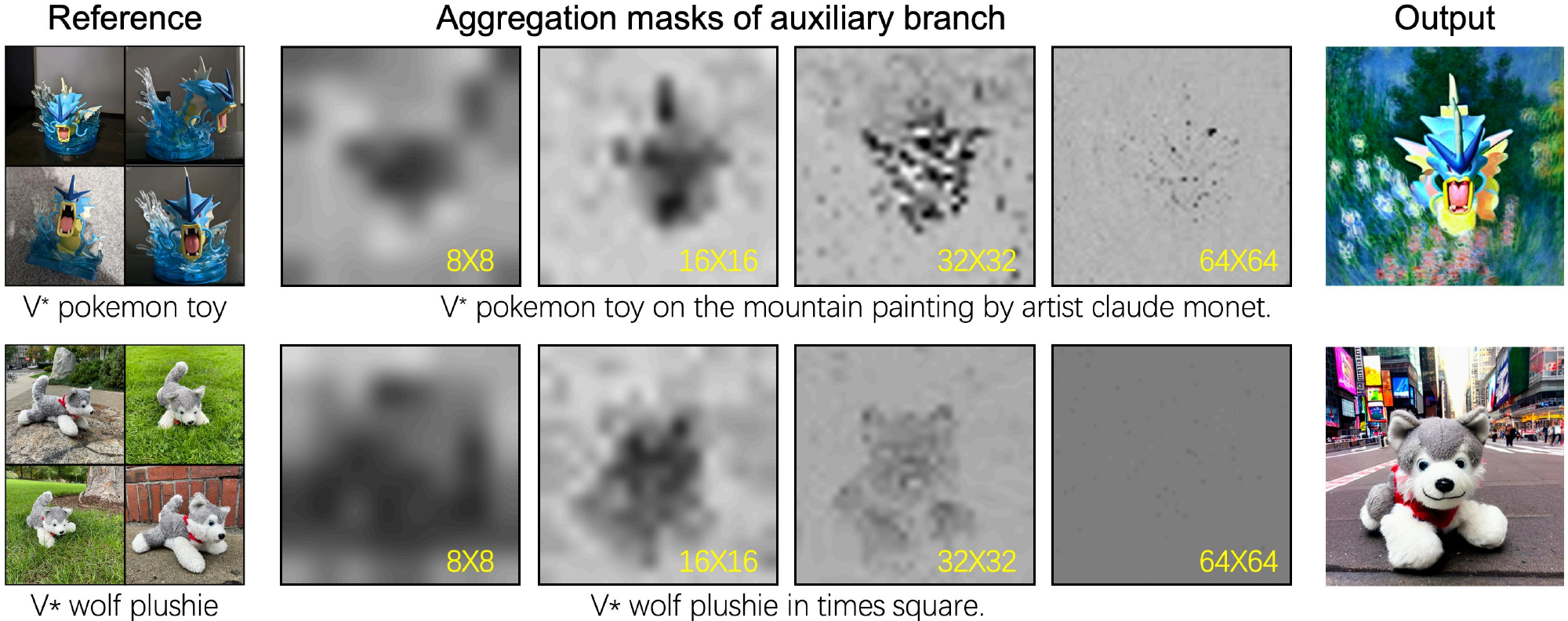}
    
    \caption{{\textbf{Visualization of aggregation masks.} 
    We visualize aggregation masks of the auxiliary branch at various scales (indicated in each image). The brighter pixels represent higher aggregation weights.
    } }
    \label{fig:vis_mask}
\end{figure}

The aggregation masks of the two branches vary spatially to enable adaptive integration.
To better explain our method, we visualize aggregation masks of two samples in~\Cref{fig:vis_mask}.
Since the masks are normalized across the two branches, we only display the masks of the auxiliary branch (\emph{i.e.}, $\hat{\mathrm{M}}_{\mathcal{B}_a}^{t,i}$ in Eqn.~\ref{eq:branch_agg}) at various scales.
The brighter pixels represent higher aggregation weights of the auxiliary branch.
As observed, different content (\emph{e.g.}, objects or style) appears to be injected at varying scales, thereby avoiding spatial confusion.
Objects are injected at small scales, while style is injected at larger scales. Objects are injected at small scales, while style is injected at larger scales.
Therefore, DCI is capable of performing prompts such as style transfer and property modification, where the functional region overlaps with the concept region.

\begin{multicols}{2}

\begin{figure}[H]
    \centering
    \includegraphics[width=0.99\linewidth]{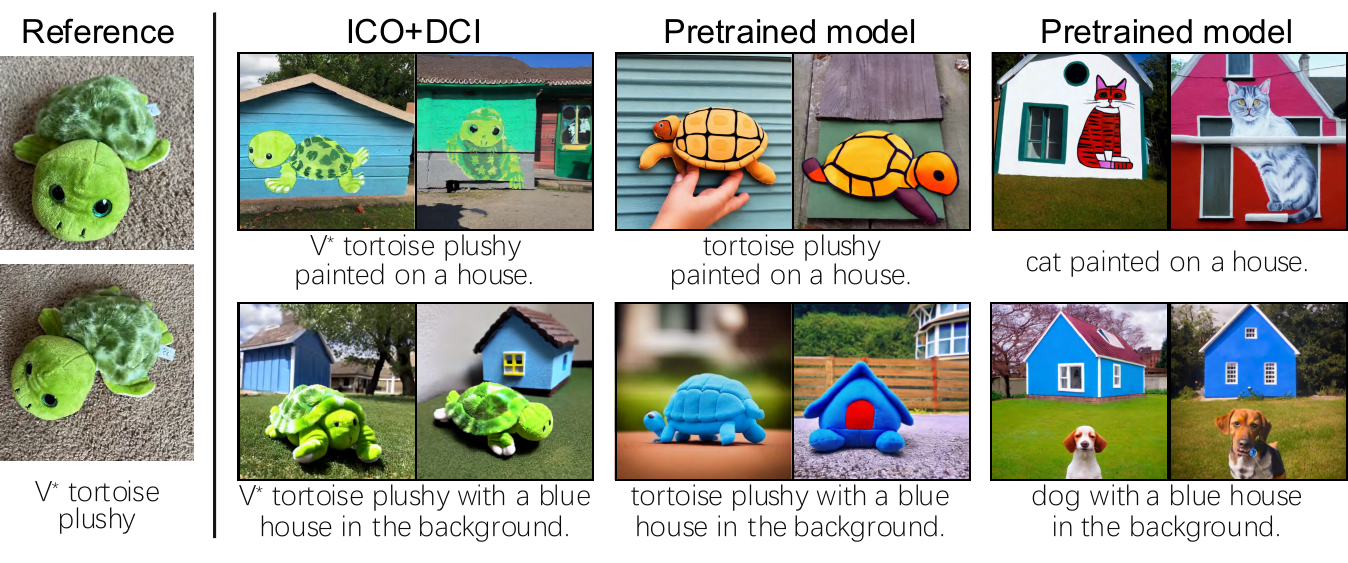}

    \caption{{\textbf{Editability enhancement by noun replacement.} 
    By selecting appropriate auxiliary nouns, DCI can successfully render the given subject in contexts where the pretrined models struggle.
    } }
    \label{fig:brk_limi}
\end{figure}
\end{multicols}

\subsection{Advances and Limitations}\label{sec:brk_limi}

A common noun that just differs from the original noun works well as the auxiliary noun in most cases. We preset common nouns for automatic auxiliary prompt construction.
Furthermore, by selecting appropriate auxiliary nouns (\emph{i.e.}, nouns that have a high probability of occurrence within the given context), DCI can overcome some limitations exhibited by pretrained models and achieve higher editability on the given concept.
As exemplified in~\Cref{fig:brk_limi}, {\menlo "cat"} has a higher model prior than {\menlo "tortoise plushy"} in the context {\menlo "painted on a house"}.
By selecting {\menlo "cat"} as the auxiliary noun, DCI successfully renders {\menlo "V* tortoise plushy"} in such a context, which even pretrained models struggle with.
Here, {\menlo "cat"} serves as an example to represent a large group of common nouns.
However, for some counterfactual or rarely occurred contexts, such as {\menlo "swimming on the moon"}, pretrained models have low model priors for all available nouns. In such cases, DCI fails to generate corresponding images faithfully.

\section{Conclusion}
In this paper, we explore and address the inherent fidelity-editability trade-off in customized generation.
We propose a \emph{``Divide, Conquer, then Integrate''} (DCI) framework, along with an \emph{Image-specific Context Optimization} (ICO) fine-tuning strategy.
ICO rectifies the customization process to obtain superior fine-tuned models, while DCI performs surgical adjustment in the early stage of denoising to liberate the fine-tuned models from the constraint of the fidelity-editability trade-off at inference.
The synergy of ICO and DCI effectively reconciles the fidelity-editability trade-off, particularly in generations with low model priors, generating customized images with high concept fidelity while faithfully adhering to the query prompt.
Furthermore, our method is capable of rendering concepts even in novel contexts where pre-trained models face challenges.
Our future work will explore more controllable customized generation conditioned on additional information such as layout~\cite{yang2023reco,feng2024layoutgpt} and depth~\cite{yan2022multi,yan2022rignet,yan2024tri}.

\section*{Acknowledgements}
The authors would like to thank the area chair and the anonymous reviewers
for their critical and constructive comments and suggestions. This work was supported by the National Natural Science Foundation of China (Grant Nos. 62361166670, 62276132 and 61876085).


%
%
\bibliographystyle{splncs04}
\bibliography{main}
\end{document}